\documentclass{article}

\PassOptionsToPackage{numbers, compress}{natbib}

\usepackage[preprint]{neurips_data_2024}

\usepackage[utf8]{inputenc} 
\usepackage[T1]{fontenc}    
\usepackage{hyperref}       
\usepackage{url}            
\usepackage{booktabs}       
\usepackage{amsfonts}       
\usepackage{nicefrac}       
\usepackage{microtype}      
\usepackage{xcolor}         
\usepackage{multirow}
\usepackage{multicol}
\usepackage{microtype}
\usepackage{hyperref}
\usepackage{url}
\usepackage{booktabs}
\usepackage{graphicx}
\usepackage{tcolorbox}
\usepackage{tablefootnote}
\usepackage{soul}
\usepackage{pifont}

\title{Are Large Language Models True Healthcare Jacks-of-All-Trades? \\Benchmarking Across Health Professions Beyond Physician Exams}

\author{%
  Zheheng Luo\thanks{equal contribution}, Chenhan Yuan\footnotemark[1], Qianqian Xie, Sophia Ananiadou \\
  Department of Computer Science\\
  National Centre for Text Mining,
  The University of Manchester\\
  Manchester, UK, M13 9PL \\
  \texttt{\{zheheng.luo, chenhan.yuan,sophia.ananiadou\}@manchester.ac.uk}\\ 
  \texttt{xqq.sincere@gmail.com}\\
}

\begin{document}

\maketitle

\begin{abstract}
Recent advancements in Large Language Models (LLMs) have demonstrated their potential in delivering accurate answers to questions about world knowledge. Despite this, existing benchmarks for evaluating LLMs in healthcare predominantly focus on medical doctors, leaving other critical healthcare professions underrepresented. To fill this research gap, we introduce the Examinations for Medical Personnel in Chinese (EMPEC), a pioneering large-scale healthcare knowledge benchmark in traditional Chinese. EMPEC consists of 157,803 exam questions across 124 subjects and 20 healthcare professions, including underrepresented occupations like Optometrists and Audiologists. Each question is tagged with its release time and source, ensuring relevance and authenticity. We conducted extensive experiments on 17 LLMs, including proprietary, open-source models, general domain models and medical specific models, evaluating their performance under various settings.
Our findings reveal that while leading models like GPT-4 achieve over 75\% accuracy, they still struggle with specialized fields and alternative medicine. Surprisingly, general-purpose LLMs outperformed medical-specific models, and incorporating EMPEC's training data significantly enhanced performance. Additionally, the results on questions released after the models' training cutoff date were consistent with overall performance trends, suggesting that the models' performance on the test set can predict their effectiveness in addressing unseen healthcare-related queries. The transition from traditional to simplified Chinese characters had a negligible impact on model performance, indicating robust linguistic versatility. Our study underscores the importance of expanding benchmarks to cover a broader range of healthcare professions to better assess the applicability of LLMs in real-world healthcare scenarios. We release the dataset and evaluation toolkit to facilitate future endeavours in developing LLMs in the healthcare domain: \url{https://github.com/zhehengluoK/eval_empec}.


\end{abstract}
\section{Introduction}

Recent advancements in Large Language Models (LLMs) have demonstrated the potential of LLM-based Artificial Intelligence (AI) in providing accurate answers to questions about world knowledge. These advancements are reflected in a series of studies and models, including but not limited to GPT-4, Gemini, Mistral, and Llama series~\citep{Achiam2023GPT4TR, Anil2023GeminiAF, Jiang2023Mistral7, Touvron2023Llama2O}. To benchmark the internal knowledge of LLMs, multiple datasets~\citep{hendrycks2020measuring, Clark2018ThinkYH, Lin2021TruthfulQAMH} have been introduced, focusing on their ability to measure and respond with accurate information. More recently, there has been a significant push towards adapting LLMs for use in the biomedicine domain~\citep{singhal2023large, chen2023huatuogptii, xie2024me}, exploring the possibility of deploying these models in real healthcare scenarios. 
To assess the effectiveness of LLMs in healthcare, considerable effort~\citep{Cai2023MedBenchAL, Wang2023CMBAC, jin2021disease,kasai2023evaluating,Pal2022MedMCQAA} has been invested in benchmarking their capabilities. Among these efforts, several studies~\citep{Cai2023MedBenchAL, Wang2023CMBAC} have incorporated long-form diagnostic questions into their benchmarks to more evaluate the models' capacities to function akin to real physicians. 
In another line of work, multiple-choice questions have emerged as a straightforward and objective means of evaluation~\citep{jin2021disease, kasai2023evaluating, Pal2022MedMCQAA}. These studies mostly utilize questions from medical licensing exams, research papers, and textbooks to assess LLMs' knowledge and suitability to serve in a physician-like role. 

\paragraph{Existing medical benchmarks are limited in scope and authenticity}
In real-world healthcare environments, medical doctors, while integral, represent only a fraction of the entire healthcare system. The International Standard Classification of Occupations (ISCO) by the International Labour Organization categorizes health professionals into two Sub-major Groups, with medical doctors listed alongside ten other minor occupational groups. These include Nursing and Midwifery Professionals and Paramedical Practitioners, among others. Each minor group comprises at least two unit groups, which in turn consist of multiple professions. This diversity in healthcare roles highlights that the expertise of medical doctors is just a subset of the broader healthcare knowledge spectrum.
Nevertheless, as shown in Table \ref{tab:vsothers}, existing medicine-related benchmarks face challenges in diversity. For instance, MedQA~\citep{jin2021disease} and MedBench~\citep{Cai2023MedBenchAL} contain questions solely for physicians. MedMCQA~\citep{Pal2022MedMCQAA} gathers questions for postgraduate medical students. CMB~\citep{Wang2023CMBAC} and CMExam~\citep{liu2024benchmarking} have made limited attempts to expand coverage beyond physicians to a maximum of five professions, leaving many health professionals unrepresented. Consequently, there remains a significant gap in our ability to assess the performance of LLMs across the broader spectrum of medical contexts.
As a result, there remains a significant gap in our ability to assess the performance of LLMs across the broader spectrum of medical contexts.
To more comprehensively gauge the effectiveness and applicability of LLMs in healthcare, it is crucial to expand the scope of benchmarks to include a wider range of professions within the healthcare system. Furthermore, despite being collected from credible sources, Table \ref{tab:vsothers} reveals that most existing works cannot reference an authoritative source for their data, casting doubt on the authenticity of the collected problems. Additionally, essential metadata, such as release dates, are often missing. Therefore, the results from existing benchmarks fail to accurately reflect LLMs' performance across the entire healthcare system.

\begin{table}[!t]
    \centering
    \small
    \caption{Review of existing healthcare-related benchmarks, \ding{51} represents the dataset that has the feature and \ding{55}  represents it does not. Source-verifiable means the dataset can be verified from its acclaimed source. Trad and simp are short for traditional and simplified respectively.}
    \begin{tabular}{lllccc}
    \toprule
         Dataset&Resources&Language&  \#Professions &\#Question&Source-verifiable \\ \midrule
         CMExam&Gov Publication&Simp Chinese&5&68,119&\ding{55}\\
         CMB&Open database&Simp Chinese&4&280,839& \ding{55}\\
         MedQA&Text Books&English, Chinese&1&61,097&\ding{55}\\
         MedMCQA&Open website\&Books &English&-&193,155&\ding{51}\\
         MedBench&Gov Publication&Simp Chinese&1&40,041&\ding{55}\\
         EMPEC&Gov Publication&Trad Chinese&
        \textbf{20}&157,803& \ding{51} \\
    \bottomrule
    \end{tabular}
    
    \label{tab:vsothers}
\end{table}

In response to the identified research gap, we introduce EMPEC, the pioneering large-scale healthcare knowledge benchmark in Chinese. EMPEC comprises 157,803 exam questions spanning 124 subjects across \textbf{20} healthcare professions. This comprehensive benchmark goes beyond the commonly assessed professions such as Physicians and Nurses, to include occupations like Optometrist and Audiologist often overlooked in previous assessments, thereby filling a critical void in existing benchmarks.
EMPEC stands out not only for its substantial size and extensive coverage but also for its authoritative and time-sensitive features. Each question within EMPEC is tagged with its release time, ensuring that the benchmark continuously integrates the latest questions from the source as soon as they are made available. This feature effectively mitigates the risk of data contamination~\citep{Nori2023CapabilitiesOG} during evaluation processes, thereby maintaining the benchmark's relevance and accuracy over time. Furthermore, the provenance of every question in EMPEC is documented, with each item linked to its original release. This verification ensures the benchmark's reliability and authenticity, positioning it as a critical tool for the evaluation of LLMs within the healthcare sector. 

We conducted extensive experiments on 17 LLMs, including 2 proprietary models and 14 open-source models, evaluating their zero-shot performance. Additionally, we include 1 fine-tuned model using the training data from the EMPEC dataset. Our analysis contrasted the performance of medical domain LLMs with general LLMs, as well as models primarily trained on Chinese versus English data. Furthermore, we evaluated the models on both the full test set and a subset comprising the most recent questions, which could be reliably assumed to be absent from the models' training data. Finally, we compared the models' performance on questions in traditional Chinese with their performance on simplified Chinese versions of the same questions. Our findings on the experimental results are as follows:


\textbf{I)} GPT-4 leads the evaluation by achieving more than 75\% accuracy while open-source LLMs are catching up with the frontier.
\textbf{II)} While leading LLMs can perform well in frequently encountered professions like Physician and Nurse,
EMEPC shows that they struggle with more specialized fields knowledge such as Dentists and Optometrists, and alternative medicine like Traditional Chinese Medicine Practitioners, underscoring the imperative for efforts towards enhancing the comprehensive healthcare capabilities of LLMs.
\textbf{III)} LLMs specifically tuned for the medical domain unexpectedly show inferior performance compared to their general-purpose counterparts. Moreover, incorporating EMPEC data into training significantly boosts model performance.
\textbf{IV)} The results on questions released later than the time cutoff of several models reveal consistency with the overall performance trends observed in the EMPEC test set, suggesting that the models' performance on the test set can be extrapolated to predict their effectiveness in addressing unseen healthcare-related queries.
\textbf{V)} The conversion from traditional to simplified Chinese characters appears to have a negligible impact on the performance of the models. This finding suggests a level of linguistic robustness in the models that could facilitate broader applicability across diverse Chinese linguistic environments.

\section{Related work}

\paragraph{Healthcare Knowledge Benchmark} 
Advancement in LLMs has demonstrated the potential of AI systems for understanding medical questions~\citep{harris2023large}, diagnosing patients~\citep{singhal2023towards}, and facilitating medical education~\citep{ravi2023large}. Multiple endeavours have been made to evaluate the medical knowledge in LLMs to propose question-answering(QA) datasets. Long-form QA datasets such as MedicationQA~\citep{abacha2019bridging}, BioASQ~\citep{krithara2023bioasq} are usually drafted and annotated by domain experts to assess LLMs via comparing their free-format response to the human written answers. 
Medical exams have become an ideal source for collecting medical QA materials due to their endorsement by national institutions and under rigorous scrutiny, Several benchmarks including MedQA~\citep{jin2021disease}, CMExam~\citep{liu2024benchmarking}, IgakuQA~\citep{kasai2023evaluating}, and Polish MFE~\citep{rosol2023evaluation} leverage medical exams from different nations to test the knowledge mostly focusing on physicians for the according languages. MedBench~\citep{Cai2023MedBenchAL} went beyond Chinese MLE to advanced physician exams such as the Resident Standardization Training Examination, the Doctor In-Charge Qualification Examination, and real-world clinic cases encompassing examinations. CMB~\citep{Wang2023CMBAC} expand the assessment of medical knowledge by bringing in exams for Nurses, Technicians, and Pharmacists.
Though of high quality, these existing benchmarks mostly focus on assessing a small proportion of health professions represented by physicians, leaving the knowledge mastered by other healthcare personnel unchecked, limiting these datasets' capability to comprehensively assess LLM's knowledge in the entire healthcare system. Our proposed EMPEC differentiate itself from previous effects by incorporating exams for in total of 20 professions in healthcare such as medical technologists, dietitians, and psychologists, whose expertise cannot be assessed by medical licensing exam questions.

\paragraph{Knowledge-related Benchmarks for LLM}
Given the advance of LLMs, there was a multitude of works focusing on benchmarking the knowledge within the models. MMLU~\citep{hendrycks2020measuring} and ARC~\citep{Clark2018ThinkYH} collected various questions from a diverse set of subjects aiming to comprehensively assess models' world knowledge. 
GSM8K~\citep{Cobbe2021TrainingVT} collected grade school math word problems to benchmark models' mathematics reasoning ability. FinBen~\citep{Xie2024TheFA} aggregated 35 datasets across 23 financial tasks aiming to fully assess LLMs' knowledge and capacity in finance. LawBench~\citep{Fei2023LawBenchBL} compiled datasets to evaluate LLM's legal capacities from three cognitive levels. However, in healthcare, most existing benchmarks concentrate on examining physicians' expertise, being unable to assess models' knowledge in the whole healthcare system.

\begin{figure}[!t]
    \centering
    \includegraphics[scale=0.45]{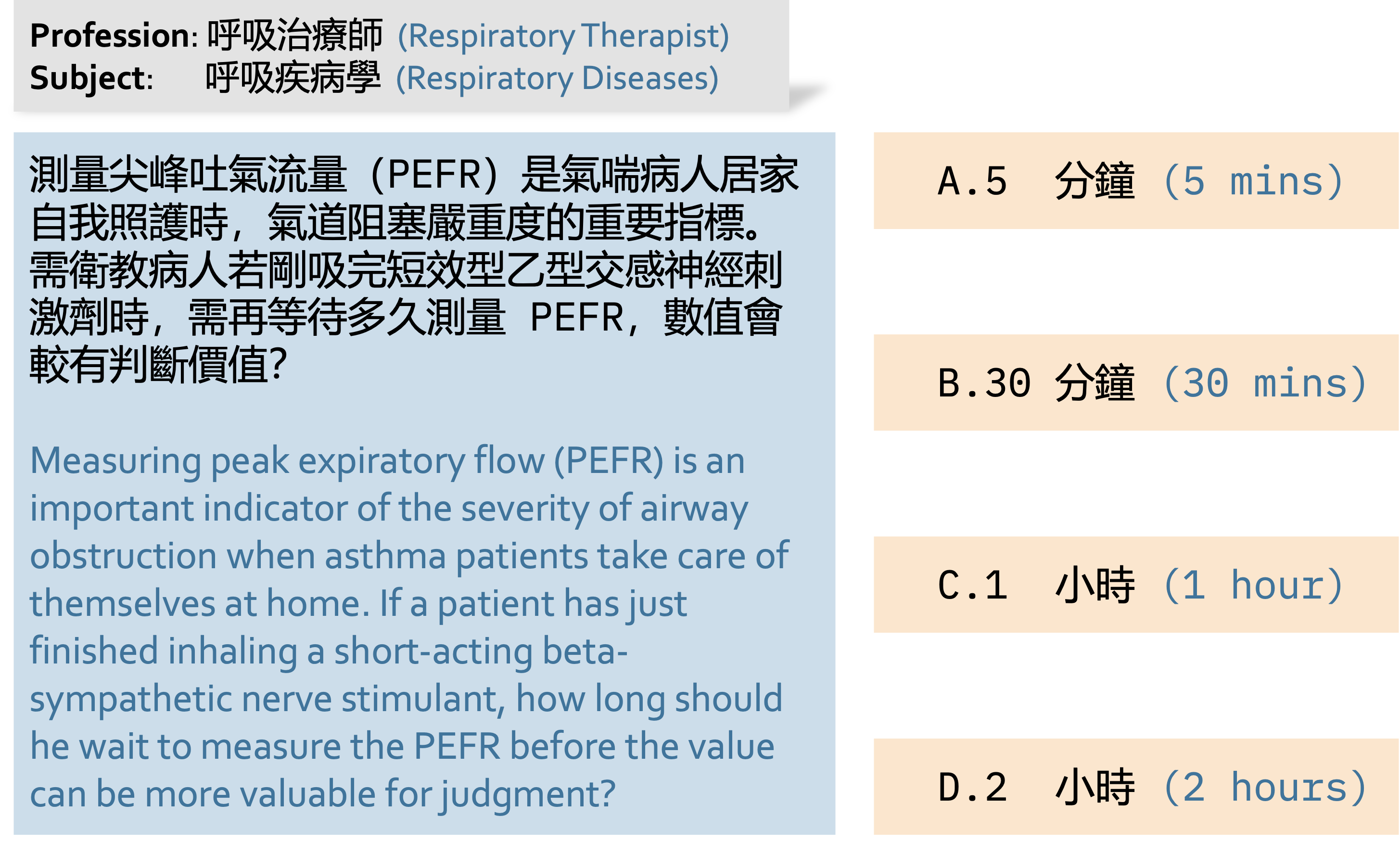}
    \caption{An example of questions in EMPEC, texts in blue are English translations of the original Chinese question and answers.}
    \label{fig:QAexample}
\end{figure}

\section{The EMPEC Dataset}
\paragraph{Data Collection and Pre-processing} EMPEC compiles officially published past exams for healthcare professionals in the Professional and Technical Examinations of Taiwan, Republic of China\footnote{\url{https://wwwq.moex.gov.tw/exam/wFrmExamQandASearch.aspx}}. Each profession's exams consist of multiple related subjects designed to comprehensively assess the candidates medical knowledge and clinical skills for the profession. Tests for some professions are conducted biannually, and we have gathered exams held from 2011 to 2024 of issuance to streamline training and evaluation processes. To ensure the quality of EMPEC, we conducted the following pre-processing steps: 1) Excluding exams for questions requiring non-textual information such as images or tables; 2) For questions that shared the same premise, we added the premise for the all following questions. 3) We remove questions belonging to subjects like "Traditional Chinese literature" and "Pharmaceutical Administration and Regulations" to rule out problems that are irrelevant or only applied in the local region to calibrate EMPEC's concentration in healthcare knowledge. 4) A MinHash deduplicate approach was applied to filter out questions that are too similar to each other. As a result, we have curated a collection of 157,803 multiple-choice questions. Originally written in traditional Chinese, we have converted the characters into simplified Chinese and provide both versions. This extensive collection covers 20 medical professions across 124 subjects. An example of questions in EMPEC is shown in Figure \ref{fig:QAexample}.

\begin{figure}[!t]
    \centering
    \includegraphics[scale=0.32]{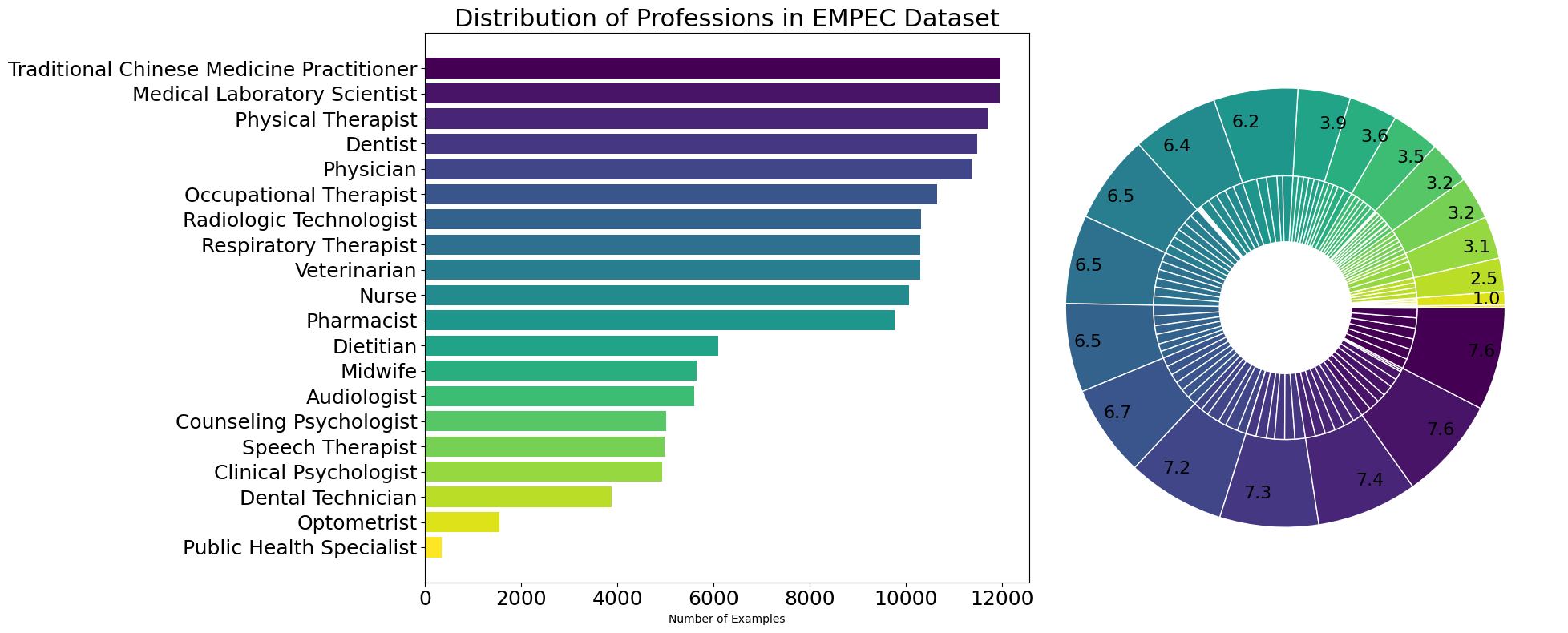}
    \caption{Distribution of professions in the EMPEC dataset. The left panel illustrates the total number of questions attributed to each healthcare profession within the dataset. The right panel provides a visual representation of the proportionate distribution of questions across the various professions.}
    \label{fig:prof_dist}
\end{figure}

\paragraph{Dataset Statistics}
In EMPEC, each question presents four options, with only one being the correct choice. Figure \ref{fig:prof_dist} illustrates the distribution of EMPEC questions across various professions. There are 11 professions taking more than 6\% per cent, 7 professions taking about 3\%. The questions for Optometrists and Public Health Specialists take less than 1\% as the exams for the two professions were recently introduced into the national exams. In Appendix \ref{appd: subj dist}, we explicitly show the subjects examined and the number of questions in each profession. The average subject of examination for each profession is more than 6, showing the diverse assessment of the expertise of each profession. Thus, EMPEC offers a more comprehensive coverage compared to prior studies~\citep{Wang2023CMBAC, jin2021disease, Cai2023MedBenchAL} which primarily focused on physicians. We split the dataset into 3 subsets, train, validation, and test. We first split the test set from the full dataset via stratified sampling by profession to ensure the same distribution. Then, we sampled 10 questions per profession from the rest data to form the validation set to facilitate few-shot evaluation. The remain data serves as the training subset and the statistics are shown in Table. \ref{tab:statistics}.
\begin{table}[!h]
    \centering
    \caption{Statistics of the each of EMPEC.}
    \begin{tabular}{ccccc}
    \toprule
         Split& \#Profession& \#Subject& \#Year &\#Question \\ \midrule
         Train&20&124&14&149,603\\
         Validation&20&100&14&200\\
         Test&20&124&14&8,000 \\
    \bottomrule
    \end{tabular}
    
    \label{tab:statistics}
\end{table}
\paragraph{Dataset Characteristics}
EMPEC has several advantages over existing medical QA benchmarks on several aspects: 1) Comprehensive evaluation for knowledge in healthcare: Unlike benchmarks such as MedBench~\citep{Cai2023MedBenchAL} and CMExam~\citep{liu2024benchmarking} which mostly focus on knowledge of physicians, EMPEC provides a thorough evaluation of knowledge and expertise of 20 healthcare professions. A review of healthcare related benchmarks is shown in Table \ref{tab:vsothers}. EMPEC covers 20 professions while none of the existing benchmarks contains more than 5 occupations.  The broad coverage enables EMPEC to serve as a valuable supplement to existing LLM benchmarks for evaluating LLMs in the healthcare domain and testing them for rare professions.
2) Extensive Question Pool: With over 157K questions from exams of 124 subjects, EMPEC surpasses the question volume of current benchmarks like CMExam, MedQA, and MedBench. This extensive question bank enhances the depth and breadth of the assessment.
3) High-Quality Assurance: Questions in EMPEC are meticulously crafted by experts and issued by the Ministry of Examination of the Republic of China, Taiwan. The questions undergo rigorous scrutiny by both the public and the medical community, ensuring their quality and authenticity. 
4) Timestamp and self-growing: every question in EMPEC is with their released year and EMPEC can automatically integrate the latest released questions. By setting the cut-off of the training data and leaving the rest of the data for evaluation, EMPEC can be used to assess if the LLMs' performances are from data contamination during training or if the knowledge has been actually embedded into the parameters.

\section{Benchmark}
\subsection{Tested Models}
\paragraph{General LLMs}
We select leading proprietary LLMs GPT-4-turbo and GPT-3.5-turbo aiming to set an upper bound for LLMs' performance on EMPEC.
For open-source models, 
We choose leading English-majored LLMs including Llama-3~\citep{llama3modelcard}, Mistral~\citep{Jiang2023Mistral7}
as well as five Chinese-majored LLMs Yi~\citep{Young2024YiOF}, Qwen1.5~\citep{qwen}, Baichuan2~\citep{Yang2023Baichuan2O}, InternLM2~\citep{cai2024internlm2}, and Ziya-Llama\citep{Fengshenbang-LM}. Both Yi and Qwen1.5 have shown performance on par with or better than GPT-4 on multiple Chinese benchmarks. 

\paragraph{Medical Domain LLMs}
As further training LLMs on in-domain data usually brings improvement in domain-specific tasks~\citep{Wang2023HuaTuoTL, Chen2023MEDITRON70BSM},
we further examined the LLMs that have been trained or fine-tuned on biomedical texts. 

\noindent HuatuoGPT2~\citep{chen2023huatuogptii}. The 13B and 34B versions of HuatuoGPT2 are pre-trained on Baichuan2-13B and Yi-34B respectively using more than 5 million synthetic medical instruction tuning data. 
HuatuoGPT2-13B is reported to outperform GPT3.5 by more than 20\% on two Chinese medical benchmarks CMB~\citep{Wang2023CMBAC}
and CMExam~\citep{liu2024benchmarking}. MMedLM2~\citep{qiu2024towards}, based on InternLM2, is further pre-trained on 22.5 billion tokens of health-related text across 6 languages including Chinese. Experimental results show that when finetuned on the same data, MMedLM2 outperform InternLM2 on medical question-answering for all six languages. MedGPT~\citep{MedicalGPT} is based on Ziya-Llama-13B~\citep{Fengshenbang-LM}, which is a Llama variant with an expanded vocabulary and further pre-trained on Chinese and English texts. MedGPT, on top of Ziya-Llama-13B, is fine-tuned on 240 million Chinese and English medical instruction-tuning data. Qwen1.5-7B-SFT, we fine-tuned a Qwen1.5-7B using the training data of EMPEC. 
Specifically, we contrast the medical LLMs with their general counterparts like Baichuan2 and Yi for HuatuoGPT2, ZiyaLlama for MedGPT, InternLM2 for MMedLM2, Qwen1.5-Chat for Qwen1.5-SFT to explore the effect of the domain-adaption training of these models.

\begin{table}[]
    \setlength\tabcolsep{2pt}
    \tiny
    \centering
     \caption{Accuracy of each model on the test set of EMPEC, split by profession. C stands for Chat version and I stands for Instruction version. Bch2 is short for Baichuan2. Some of the profession names are abbreviated; please refer to Appendix \ref{app:abbr} for their full corresponding titles.}
    \begin{tabular}{p{0.1\linewidth}ccccccccccccccccc}
    \toprule
         \multirow{2}{*}{Split}&&&&&&&&&&Accuracy$\ast$&&&&\\  \cline{2-18}

&GPT4&GPT3.5&\multicolumn{3}{c}{Qwen1.5}&\multicolumn{2}{c}{Llama3-I}&\multicolumn{2}{c}{Yi-C}&\multicolumn{2}{c}{HuatuoGPT2}&Bch2-C&Mistral-I&InternLM2&MMedLM2&MedGPT&Ziya\\
&-&-&70B-C&7B-C&7B-SFT&70B&8B&34B&6B&34B&13B&13B&7B&7B&7B&7B&7B\\ \hline
Nurse&\textbf{82.75}&63.53&77.65&61.18&69.22&70.20&52.75&73.53&48.04&41.96&17.25&55.10&40.20&62.16&58.43&24.31&29.41\\
Dentist&\textbf{69.27}&49.83&53.30&47.22&53.99&57.47&49.31&56.60&34.38&33.85&18.40&42.71&35.24&46.18&47.22&24.65&23.61\\
Midwife&\textbf{85.31}&58.04&69.23&49.65&62.24&72.73&53.85&69.93&39.16&41.61&17.48&49.30&36.71&55.24&51.05&23.08&23.43\\
Physician&\textbf{89.90}&67.42&72.30&58.01&68.47&83.62&68.29&73.87&46.34&42.86&14.81&54.70&45.12&62.89&62.37&25.09&25.78\\
Dietitian&\textbf{74.76}&53.40&70.55&52.75&61.49&65.37&47.90&63.43&39.16&44.66&20.71&50.49&37.54&56.96&52.43&29.77&20.71\\
Pharmacist&\textbf{75.86}&58.15&60.36&49.09&57.14&72.03&53.32&64.59&42.05&45.88&15.90&45.07&39.44&56.74&51.71&30.38&26.76\\
Audiologist&\textbf{72.18}&53.17&63.03&44.37&54.23&59.51&44.72&57.39&34.51&34.15&19.37&40.14&33.45&49.30&44.37&23.59&20.77\\
Optometrist&\textbf{64.10}&51.28&51.28&44.87&53.85&52.56&46.15&48.72&37.18&39.74&16.67&41.03&25.64&44.87&42.31&20.51&21.79\\
Veterinarian&\textbf{78.35}&55.56&63.22&49.81&63.41&63.22&50.38&60.73&38.70&36.02&17.62&43.87&38.12&53.07&52.49&21.26&25.29\\
Speech Therap&\textbf{69.05}&46.43&54.76&40.87&59.13&52.38&36.90&51.19&23.81&33.33&19.05&38.49&29.37&40.87&41.67&27.78&22.22\\
Dental Tech&\textbf{59.39}&41.12&58.88&50.25&53.81&46.70&44.16&52.28&36.55&35.03&13.71&38.07&23.86&44.67&43.65&31.47&26.90\\
Phys Therap&\textbf{70.25}&51.43&52.61&52.27&58.49&60.67&52.10&59.33&37.98&36.13&19.50&46.89&36.47&53.28&50.76&28.24&24.54\\
Resp Therap&\textbf{75.43}&53.74&52.59&48.56&57.01&61.61&49.52&56.62&36.85&35.32&20.54&42.42&39.16&49.90&47.22&24.38&24.38\\
Clin Psych&\textbf{87.65}&67.73&77.29&59.76&65.74&72.11&59.36&72.11&46.22&43.03&19.92&54.98&45.82&67.33&64.54&31.47&31.87\\
Occup Therap&\textbf{76.85}&58.89&60.93&53.52&65.56&62.41&51.85&61.85&39.44&36.48&20.56&50.37&40.56&56.11&54.63&30.37&26.48\\
Rad Tech&\textbf{76.97}&54.32&58.73&43.57&59.12&64.68&51.44&54.70&30.33&33.40&17.66&38.96&34.36&51.06&48.37&25.53&20.92\\
Counsel Psych&\textbf{79.30}&60.94&75.78&61.72&68.75&67.19&60.55&73.44&53.52&42.19&16.02&60.55&43.75&64.84&61.72&31.25&28.91\\
PH Spec&\textbf{82.35}&70.59&58.82&52.94&76.47&52.94&58.82&64.71&47.06&41.18&41.18&52.94&47.06&52.94&41.18&41.18&17.65\\
Med Lab Sci.&\textbf{85.17}&67.71&68.86&55.68&63.26&76.77&58.81&66.72&40.53&43.00&18.12&54.04&44.81&60.30&58.32&30.31&26.19\\
TCM Prac&50.08&36.24&\textbf{63.26}&39.37&54.53&44.81&34.27&55.85&32.13&21.91&13.84&43.16&24.71&49.59&44.98&16.31&17.30\\
Micro Avg.&\textbf{75.35}&55.66&63.24&50.79&60.84&64.46&51.41&62.29&38.79&37.45&17.81&47.20&37.44&54.50&52.08&26.08&24.51\\
Marco Avg.&\textbf{75.25}&55.96&63.17&50.77&61.27&63.02&51.23&61.90&39.18&38.06&18.86&47.17&37.09&53.94&51.02&27.00&24.26\\

\bottomrule
\multicolumn{16}{l}{$\ast$ A random guess baseline gets 24.96\% micro average accuracy}
    \end{tabular}
   
    \label{tab:acc}
\end{table}

\subsection{Evaluation settings}
We use the \textit{0125} version of GPT-turbo models. To facilitate zero-shot prompting, we choose the "Chat" or "Instruct" versions of Llama-3, Mistral(v0.2), Baichuan2, Qwen1.5, and Yi. We fine-tuned Qwen1.5-7B on the EMPEC training subset for 3 epochs using the learning rate of 1e-4 on 2 A100 GPUs. We use vLLM 0.4.1~\citep{kwon2023efficient} and enable greedy decoding to ensure the stability and reproducibility of the results. The prompt used in zero-shot prompting and fine-tuning is in Appendix \ref{app: prompt}.
Moreover, as existing benchmarks suffer the problem of data leakage~\citep{Nori2023CapabilitiesOG, zhang2024careful}, we explicitly conduct the evaluation on the question from the exams held in 2024 on models of which the pretraining data cut-off is before 2024. 
In addition, to investigate if prompting with traditional Chinese characters would make a difference from using simplified Chinese characters, we adopted \textit{zhconv}\footnote{\url{https://github.com/gumblex/zhconv}} to convert the questions into simplified Chinese.
Those questions that are filtered by the API provider or models fail to respond with the correct choice are deemed wrongly answered. A baseline is set by randomly picking a choice from A to D.


\section{Analysis}
In general, all the tested models show unsatisfying performance. The average accuracy of 12 out of the 17 tested models is under 60\% on EMPEC where random guessing can achieve about 25\%. The results suggest the difficulty of our benchmark and the healthcare knowledge gap in existing LLMs.

\subsection{Results of general LLMs}

From the accuracy of each profession and the average overall profession, GPT-4 shows an evident lead among tested models, with only a second Qwen1.5-70B-Chat in Traditional Chinese Medicine Practitioner. Another proprietary model, GPT-3.5 shows a nearly 20\% performance drop from GPT 4. Open-source models lag far behind GPT 4, the accuracy of the best-performing models, Qwen1.5-70B-Chat and Llama3-70B-Instruct, is around 63\% — 12\% less than GPT-4. However, gained from training on a large proportion of Chinese data, Yi-34B and Qwen1.5-70B beat GPT 3.5.
Observing results of Yi, Qwen1.5, and Llama-3, we see that increasing model size can benefit models' performance as expected. Among models of size around 7B, InternLM2, Qwen1.5-Chat, and Llama3-8B-Instruct achieve over 50\% accuracy while Mistral and Yi are slightly under 40\%.


\subsection{Results of medical domain LLMs}
HuatuoGPT2, MMedLM2, and MedGPT, despite of being based on Chinese-focused LLMs and specifically trained on medical data, show poor performance on EMPEC. The results of MedGPT  are slightly higher than random guess while HuatuoGPT2-13B even under-performs the random baseline. MMedLM2 performs best among medical LLMs, achieves 45\% accuracy with 7B parameters, but still lags behind Yi-6B and Qwen1.5-7B. We noticed that part of why the poor performance is due to loss of instruction-following ability. For instance, HuatuoGPT2, despite of being able to generate plausible response, sometimes deviates from the questions and often fail to conclude an answer. HuatuoGPT2 underperforms Baichuan-2-13B-Chat and Yi-34B-Chat which shares the same base model but fine-tuning on general instruction data. 
The findings are different from the results reported in~\citet{Wang2023CMBAC} where HuatuoGPT2 outperforms Baichuan2 by a large margin on Chinese medical questions. Also, InternLM2 leads MMedLM2 by more than 2\% accuracy while fine-tuned MMedLM2 outperforms InternLM2 on MedQA~\citep{jin2021disease}. Only MedicalGPT outperforms Ziya-Llama by around 2\%. In conclusion, we do not observe evidence enhancement on EMPEC brought by fine-tuning or pre-training in the three works, contrasting the findings on existing benchmarks like CMB~\citep{Wang2023CMBAC}. Therefore, we argue that, compared to the existing Chinese healthcare benchmark, our new EMPEC provides a more robust platform for the evaluation of domain-adaptation LLMs. This finding further suggests that the current improvement in domain adaptation could result in an over-fit of the tested distribution, which might not hold once the distribution switches.
Moreover, our fine-tuned Qwen1.5-7B model achieves 61\% accuracy, nearly 11\% better than the Chat counterpart and close to Qwen1.5-70B-Chat, suggesting that the training data of EMPEC can be an effective supplement of current medical domain adaptation endeavours.


\subsection{Performance by profession}
We further examined the performance of the models by profession. While GPT-4 prevails on 19 of the 20 professions, Qwen1.5-70B-Chat achieves the highest accuracy in Traditional Chinese Medicine Practitioner.
Then to inspect which professions are the best and worst performed, we rank the accuracy of each profession within every model and then aggregate rankings across models. The best three performed professions are Clinical Psychologist, Physician, and Nurse while the models mostly struggle to answer questions for Dental technician, Speech Therapist, Optometrist, Dentist, and TCM Practitioner, which seem connect more to specific medical subject. We assume this difference is rooted in the frequency of knowledge in the training data of these models and the rareness of knowledge of these bad-performing professions as \citet{kandpal2023large} has suggested. Take GPT-4 as an example, it achieves nearly 90\% accuracy for physicians of which related-documents are rich on the Internet and books, while its performance on rarely touched professions like Dental technician and Traditional Chinese Medicine Practitioner is lower than 60\%.
Interestingly, all tested medical domain LLMs except Qwen1.5-7B-SFT do not show a better performance compared to their general counterparts even on physicians which is the focus of their finetuning data. However they generally unsatisfying performance in each profession indicating the importance of data diversity in building an LLM for healthcare.
By breaking the overall results into professions, EMPEC is able to precisely detect LLM's knowledge gap in the healthcare domain, and we want to raise attention to creating models that could also achieve good performance in rarely seen professions.
\begin{figure}[!t]
    \centering
    \includegraphics[scale=0.5]{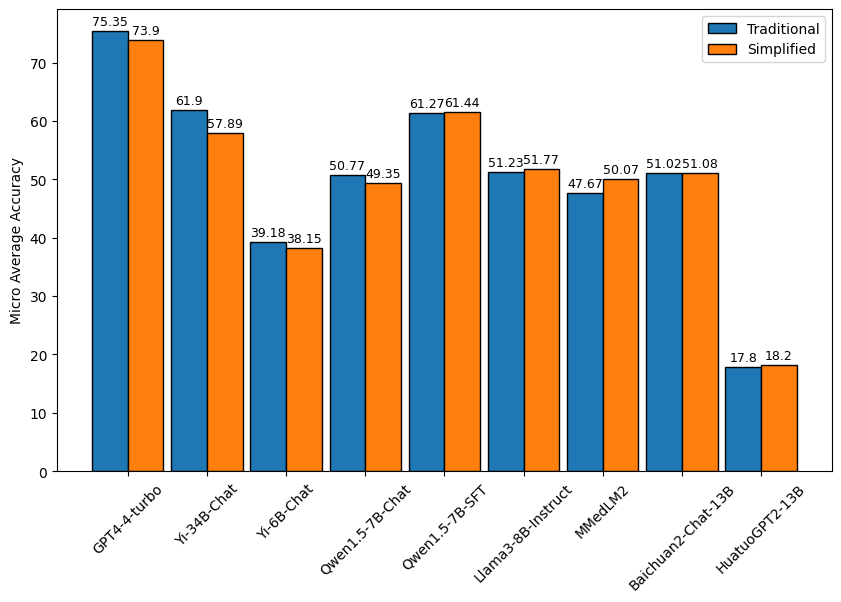}
    \caption{Performance of models on traditional Chinese and simplified Chinese.}
    \label{fig:tradvssimp}
\end{figure}
 
\subsection{Results on the most recent questions}
During the collection of EMPEC, we found some of the past questions are freely available on the Internet, the possibility of the data having been seen by LLMs can not be ruled out. Thus, we explicitly composed a dataset containing 3497 questions issued in the exams in 2024, which should not be seen by the models of which cut-offs are before 2024. The statistics of this dataset are in Table \ref{tab:q2024statistics}. We tested GPT-4-turbo(cutoff on Dec. 2023), GPT-3.5-turbo(cutoff on Sept. 2021), Yi (cutoff on Dec. 2023), Baichuan-2 (released on Sept. 2023), and Mistral (released at Dec. 2023) on the questions released in 2024. It is important to note that this split might have overlapped with the test set of EMPEC and the included professions are limited as the exams for the rest professions have not been held yet.
The results are in Table \ref{tab:2024}.
In general, we observed the results of questions 2024 are very close to the results obtained from the test set across assessed models. In other words, the performance rankings are held as in the test set. Moreover, the professions where models perform well such as Physician and Nurse and perform poorly such as Dentist and Traditional Chinese Medicine Practitioner also keep in the latest questions set.  The findings strongly suggest the robustness of the EMPEC test set as the performance of models is consistent with that of the unseen data. In addition, it also indicates that the tested models do not gain their performance by reciting seen examples.

\begin{table}[b!]
    \centering
    \tiny
    \caption{Performance of models on questions from 2024. Colored numbers are the results of questions from exams in 2024 while the black ones are the results of the test set. Red indicates higher while blue suggests degrade.}
    \begin{tabular}{lcccccccc}
    \toprule
         \multirow{2}{*}{Profession} & \multicolumn{5}{c}{Accuracy}\\
&GPT 4&GPT 3.5&Yi 34B&Baichuan2-Chat-13B&Mistral-Instruct-7B-v0.2\\\hline
Phys Therap&\textcolor{red}{73.05} (70.25)&\textcolor{blue}{50.78} (51.43)&\textcolor{blue}{54.34} (59.33)&\textcolor{red}{49.00} (46.89)&\textcolor{red}{36.97} (36.47)\\
Rad Tech&\textcolor{blue}{72.58} (76.97)&\textcolor{blue}{51.52} (54.32)&\textcolor{red}{58.68} (54.70)&\textcolor{red}{44.63} (38.96)&\textcolor{red}{34.44} (34.36)\\
Physician&\textcolor{blue}{85.74} (89.90)&\textcolor{blue}{63.39} (67.42)&\textcolor{blue}{72.18} (73.87)&\textcolor{blue}{48.74} (54.70)&\textcolor{blue}{44.77} (45.12)\\
TCM Prac&\textcolor{red}{57.05} (50.08)&\textcolor{red}{36.58} (36.24)&\textcolor{blue}{50.65} (55.85)&\textcolor{red}{46.97} (43.16)&\textcolor{blue}{24.03} (24.71)\\
Dentist&\textcolor{blue}{67.22} (69.27)&\textcolor{blue}{47.29} (49.83)&\textcolor{blue}{50.82} (56.60)&\textcolor{blue}{39.76} (42.71)&\textcolor{red}{38.82} (35.24)\\
Pharmacist&\textcolor{red}{76.24} (75.86)&\textcolor{blue}{58.01} (58.15)&\textcolor{blue}{63.54} (64.59)&\textcolor{red}{47.51} (45.07)&\textcolor{blue}{38.40} (39.44)\\
Med Lab Sci.&\textcolor{red}{85.97} (85.17)&\textcolor{blue}{63.92} (67.71)&\textcolor{red}{67.93} (66.72)&\textcolor{blue}{47.88} (54.04)&\textcolor{blue}{43.21} (44.81)\\
Dietitian&\textcolor{red}{75.93} (74.76)&\textcolor{red}{55.51} (53.40)&\textcolor{blue}{60.29} (63.43)&\textcolor{blue}{45.59} (50.49)&\textcolor{red}{40.44} (37.54)\\
Nurse&\textcolor{red}{87.76} (82.75)&\textcolor{red}{71.31} (63.53)&\textcolor{red}{78.48} (73.53)&\textcolor{red}{56.54} (55.10)&\textcolor{red}{43.46} (40.20)\\
Micro Avg.&\textcolor{red}{75.12} (74.54)&\textcolor{blue}{54.48} (55.47)&\textcolor{blue}{61.11} (62.82)&\textcolor{blue}{47.07} (47.70)&\textcolor{red}{37.95} (37.44)\\

    \bottomrule
    \end{tabular}
    
    \label{tab:2024}
\end{table}

\subsection{Traditional vs Simplified Chinese}
The difference between traditional Chinese and simplified Chinese can affect the readers' analytic skills~\citep{Liu2012ThePO}. We further investigated the performance of a subset of tested models on the test set in traditional Chinese and simplified Chinese. The results are shown in Figure \ref{fig:tradvssimp}. From traditional to simplified, contrary effects are observed among the models with GPT-4, Yi-Chat, and Qwen1.5-Chat suffering slight performance drops while HuatuoGPT2, MMedLM2, Llama3-Instruct, Qwen1.5-SFT enjoying slight improvement. However, the performance across these models between the two kinds of Chinese does not vary greatly though the traditional and simplified characters do not share the same tokens. We assume that the character difference in traditional and simplified Chinese does not affect the model's performance as the co-occurrence relations of these characters are similar in the two kinds of Chinese.

\section{Conclusion and Discussions}
We introduced EMPEC, the most comprehensive healthcare knowledge benchmark to date, encompassing 157,803 questions across 124 subjects and 20 healthcare occupations. EMPEC goes beyond the physician-centric focus of previous medical benchmarks to provide a holistic assessment of LLMs’ knowledge across a wide spectrum of health disciplines.
Our extensive experiments, which tested both proprietary and open-source models, revealed several important findings. While the best general LLMs performed reasonably well on common professions such as Physicians and Nurses, they struggled with rarer specialties like Dental Technicians, Optometrists, and Traditional Chinese Medicine (TCM) Practitioners. This highlights the need for further work to enhance LLMs' mastery of healthcare knowledge beyond the physician expertise that has been the primary focus so far.
Interestingly, we found that existing medical domain-specific LLMs did not perform better on EMPEC compared to their general counterparts not specialized in healthcare. However, the performance of models trained on the EMPEC training data showed significant improvement. This suggests that current approaches to domain adaptation may be insufficient and demonstrates that the EMPEC test set can serve as a more robust benchmark. Additionally, the EMPEC training data is an effective supplement to current domain adaptation efforts.
Furthermore, our experiments indicate that models' performance on the EMPEC test set can predict their effectiveness in addressing unseen healthcare-related queries. We also found that the transition from traditional to simplified Chinese characters had negligible impact on model performance.
EMPEC lays the groundwork for developing advanced LLMs for the healthcare domain, providing a more comprehensive and nuanced evaluation of their capabilities across a broad range of health professions.
\paragraph{Limitations} EMPEC has several limitations: 1) All questions in the dataset are multiple-choice, which means that random guessing could still result in a certain degree of accuracy. 2) EMPEC is available only in Chinese, limiting its applicability for assessing large LLMs' healthcare knowledge in English, the predominant language in healthcare. 3) Some professions included in EMPEC have a relatively smaller representation compared to others, which may restrict EMPEC's effectiveness in evaluating knowledge in those specific professions.
\paragraph{Ethics Statement}
The questions included in EMPEC are officially published by the Ministry of Education, Taiwan, Republic of China. These authorities do not impose a specific license or restrict the distribution of this data.\footnote{\url{https://wwwq.moex.gov.tw/exam/wFrmExamQandASearch.aspx}} For more information, please refer to the provided link to view the original declaration. It is important to note that the EMPEC dataset is intended exclusively for academic and research purposes. Any commercial use or other forms of misuse that diverge from this intended purpose are strictly prohibited. We urge all users to adhere to this stipulation to ensure the ethical and proper use of this resource.

\paragraph{Societal Impacts} Although EMPEC is designed to improve the evaluation of LLMs in the medical field, it should not be used to assess individual medical competence or for patient diagnosis. Any conclusions drawn from models trained on this dataset should take its limitations into account. The dataset's use should be confined to research purposes to prevent potential misuse.

\bibliographystyle{plainnat}
\bibliography{refs}

\clearpage

\appendix

\section{Prompt}
\label{app: prompt}
\begin{figure}[!h]
    \centering
    \includegraphics[scale=0.5]{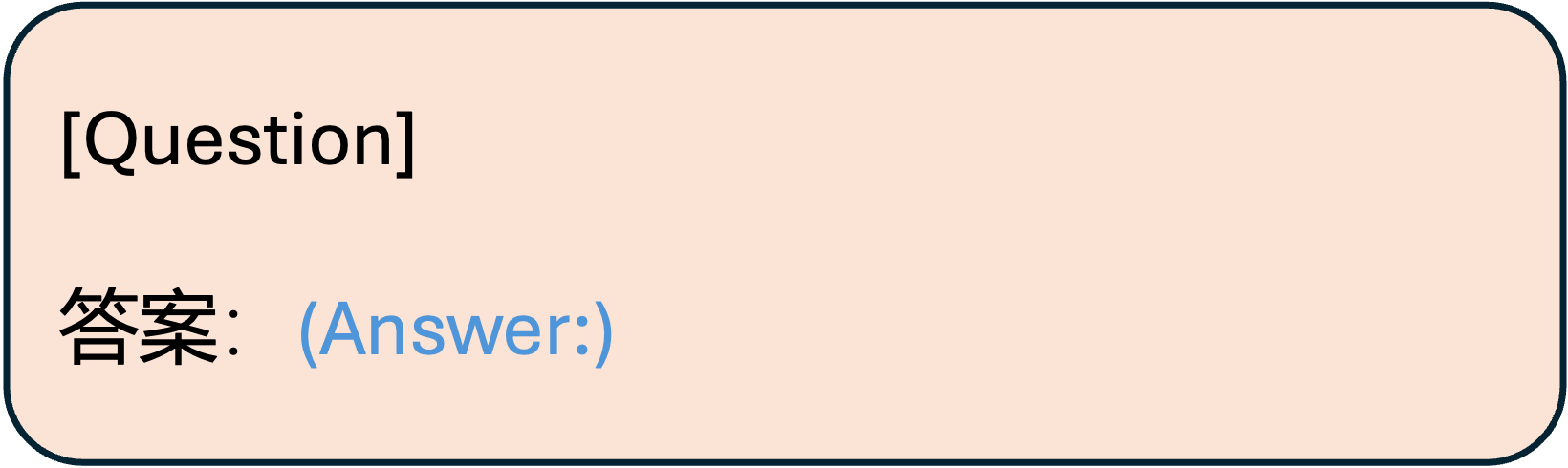}
    \caption{The prompt used in the zero-shot evaluation and supervised fine-tuning. The texts in blue are the English translations of the Chinese content.}
    \label{fig:prompt}
\end{figure}

The prompt used in the zero-shot evaluation and supervised fine-tuning is shown in Fig. \ref{fig:prompt}.

\section{Abbreviation of professions}
\label{app:abbr}
The correspondence of the abbreviated names and the full titles are shown in Table \ref{tab:abbr}.
\begin{table}[!h]
    \centering
    \begin{tabular}{cc}
    \toprule
Short Name&Profession\\  \hline

Resp Therap&Respiratory Therapist\\
Counsel Psych&Counselling Psychologist\\
Occup Therap&Occupational Therapist\\
Med Lab Sci.&Medical Laboratory Scientist\\
Phys Therap&Physical Therapist\\
Clin Psych&Clinical Psychologist\\
Rad Tech&Radiologic Technologist\\
TCM Prac&Traditional Chinese Medicine Practitioner\\
Speech Therap&Speech Therapist\\
Dental Tech&Dental Technician\\
PH Spec&Public Health Specialist\\
\bottomrule
    \end{tabular}
    \caption{Full Titles for Abbreviated Profession Names.}
    \label{tab:abbr}
\end{table}

\section{Statistics of EMPEC questions issued in 2024}

\begin{table}[]
    \centering
    \begin{tabular}{cc}
    \toprule
    Professors&Number of questions\\\hline
Physical Therapist & 449 \\
Radiologic Technologist & 363 \\
Physician & 478 \\
Traditional Chinese Medicine Practitioner & 462 \\
Dentist & 425 \\
Pharmacist & 362 \\
Medical Laboratory Scientist & 449 \\
Dietitian & 272 \\
Nurse & 237 \\
\bottomrule
    \end{tabular}
    \caption{Number of questions for each profession in questions issued in 2024}
    \label{tab:q2024statistics}
\end{table}

\section{Subject distributions in EMPEC}
\label{appd: subj dist}
We show the distribution of subjects in EMPEC in Table. \ref{tab:subjects statistics}.
\begin{table}[]
\tiny
    \centering
    \caption{Subjects examined for each profession}
    \begin{tabular}{p{0.2\linewidth}p{0.5\linewidth}cc}
\toprule
\textbf{Profession} & \textbf{Subjects} & \textbf{\#Subjects} & \textbf{\#Questions} \\
\midrule
Traditional Chinese Medicine Practitioner&Basic Chinese Medicine (1-2); Clinical Chinese Medicine (1-4); Pharmacy and Biopharmaceutics&7&11954 \\ \midrule
Medical Laboratory Scientist&Clinical Hematology and Blood Bank; Biochemistry and Clinical Biochemistry; Microbiology and Clinical Microbiology; Clinical Serum Immunology and Clinical; Medical Molecular Testing and Clinical; Clinical Physiology and Pathology; Clinical Mirror Examination&7&11938 \\ \midrule
Physical Therapist&Basic Physical Therapy; Cardiopulmonary and Pediatric Disease Therapy; Orthopedic Disease Physical Therapy; Neurological Disease Physical Therapy; Introduction to Physical Therapy; Physical Therapy Techniques&6&11698 \\ \midrule
Dentist&Dentistry (1-6)&6&11468 \\ \midrule
Physician&Medicine(1-6); Clinical Psychology Special Topics&7&11366 \\ \midrule
Occupational Therapist&Anatomy and Physiology; Psychological Disability Occupational Therapy; Occupational Therapy Techniques; Introduction to Occupational Therapy; Pediatric Occupational Therapy; Physiological Disability Occupational Therapy&6&10646 \\ \midrule
Radiologic Technologist&Basic Medical Science; Radiation Therapy Principles and Techniques; Nuclear Medicine Diagnosis Principles and Techniques; Medical Physics and Radiation Safety; Radiological Equipment; Radiological Diagnosis Principles and Techniques&6&10307 \\ \midrule
Respiratory Therapist&Cardiopulmonary Basic Medical Science; Respiratory Principles and Applications; Intensive Respiratory Therapy; Respiratory Therapy Equipment; Respiratory Diseases; Basic Respiratory Therapy&6&10301 \\ \midrule
Veterinarian&Veterinary Laboratory Diagnosis; Veterinary Pharmacology; Veterinary General Diseases; Veterinary Infectious Diseases; Veterinary Pathology; Veterinary Public Health&6&10292 \\ \midrule
Nurse&Internal and Surgical Nursing; Basic Nursing; Basic Medical Science; Mental Health and Community Health Nursing; Obstetric and Pediatric Nursing; Overview of Basic Medical Science; Obstetrics, Psychiatry and Community; Overview of Basic Nursing; Overview of Internal and Surgical Nursing&9&10066 \\ \midrule
Pharmacist&Pharmacotherapy; Pharmacy; Dispensing and Clinical Pharmacy; Pharmacology and Pharmaceutical Chemistry; Pharmacy and Biopharmaceutics&5&9767 \\ \midrule
Dietitian&Group Meal Design and Management; Diet Therapy; Nutrition; Food Hygiene and Safety; Physiology and Biochemistry; Public Health Nutrition&6&6088 \\ \midrule
Midwife&Midwifery (1-2); Nursing for All Specialties; Basic Medical Science; Basic Nursing&5&5642 \\ \midrule
Audiologist&Basic Audiology; Hearing and Language Communication Disorders; Health of Auditory and Balance Systems; Electrophysiological Audiology; Behavioral Audiology; Principles and Practice of Hearing Aids&6&5600 \\ \midrule
Counseling Psychologist&Counseling and Psychotherapy Theories; Counseling and Psychotherapy Practice and; Human Behavior and Development; Group Counseling and Psychotherapy; Case Assessment and Psychological Evaluation; Psychological Foundations of Counseling; Mental Health and Abnormal Psychology; Mental Health; Psychological Testing and Assessment; Counseling and Psychotherapy Practice&10&5014 \\ \midrule
Speech Therapist&Articulation and Fluency Disorders; Basic Linguistics; Communication Disorders Overview; Neurological Communication Disorders; Child Language Disorders; Voice and Swallowing Disorders&6&4973 \\ \midrule
Clinical Psychologist&Special Topics in Clinical Psychology (1-2); Clinical Psychology Special Topics (1-2); Basic Clinical Psychology&5&4923 \\\midrule
Dental Technician&Dental Technology (1-4)&4&3885 \\\midrule
Optometrist&Optometry; Low Vision; Eye Anatomy, Physiology and Ethics; Contact Lens and Dispensing; Visual Optics&5&1538 \\\midrule
Public Health Specialist&Health Administration and Management; Epidemiology; Environmental and Occupational Health; Biostatistics; Health Social Behavior&5&337 \\\midrule
Total && 124 & 157803 \\ \midrule
\bottomrule
    \end{tabular}
    \label{tab:subjects statistics}
\end{table}

\end{document}